\documentclass{article}
\usepackage{setspace}

\PassOptionsToPackage{numbers, sort, compress}{natbib}

\usepackage[final]{neurips_2025}

\usepackage[utf8]{inputenc} 
\usepackage[T1]{fontenc}    
\usepackage{hyperref}       
\usepackage{url}            
\usepackage{booktabs}       
\usepackage{amsfonts}       
\usepackage{nicefrac}       
\usepackage{microtype}      
\usepackage{xcolor}         
\usepackage{tcolorbox}
\usepackage{enumitem}
\usepackage{float}

\usepackage{multirow}
\usepackage{graphicx} 
\usepackage{booktabs}
\usepackage[table]{xcolor}
\usepackage{amssymb}
\usepackage{pifont}
\usepackage{graphicx}
\usepackage{multirow}
\usepackage{makecell}
\usepackage{booktabs}
\usepackage{geometry}
\usepackage{subcaption}
\usepackage{caption}
\usepackage{float}
\usepackage{array}
\usepackage{tabularx}
\newcolumntype{Y}{>{\centering\arraybackslash}X}
\definecolor{lightblue}{RGB}{224,239,251}
\definecolor{deepSkyBlue}{rgb}{0.0, 0.75, 1.0}
\definecolor{sunYellow}{rgb}{1.0, 1.0, 0.1}
\definecolor{amberOrange}{rgb}{1.0, 0.49, 0.0}  

\title{WavBench: Benchmarking Reasoning, Colloquialism, and Paralinguistics for End-to-End Spoken Dialogue Models}

\author{
  Yangzhuo Li$^{\clubsuit}$\thanks{Equal contribution.} \quad
  Shengpeng Ji$^{\spadesuit}$\footnotemark[1]
  ~\thanks{Corresponding author.} \quad
  Yifu Chen$^{\spadesuit}$\footnotemark[1] \quad
  Tianle Liang$^{\spadesuit}$\footnotemark[1] \quad
  Haorong Ying\footnotemark[1]
  \\
  \textbf{
    Yule Wang$^{\heartsuit}$ \quad
    Junbo Li \quad
    Jun Fang \quad
    Zhou Zhao$^{\spadesuit}$\footnotemark[2]
  } \\[0.5ex]
  $^{\clubsuit}$Xiamen University \quad
  $^{\spadesuit}$Zhejiang University \quad
  $^{\heartsuit}$CUHK-Shenzhen\thanks{The Chinese University of Hong Kong, Shenzhen} \\[0.5ex]
  \texttt{liyangzhuo49@gmail.com} \quad \texttt{shengpengji@zju.edu.cn}
}

\begin{document}

\maketitle

\begin{abstract}

With the rapid integration of advanced reasoning capabilities into spoken dialogue models, the field urgently demands benchmarks that transcend simple interactions to address real-world complexity. However, current evaluations predominantly adhere to text-generation standards, overlooking the unique audio-centric characteristics of paralinguistics and colloquialisms, alongside the cognitive depth required by modern agents. To bridge this gap, we introduce WavBench, a comprehensive benchmark designed to evaluate realistic conversational abilities where prior works fall short. Uniquely, WavBench establishes a tripartite framework: 1) \textbf{Pro} subset, designed to rigorously challenge reasoning-enhanced models with significantly increased difficulty; 2) \textbf{Basic} subset, defining a novel standard for spoken colloquialism that prioritizes "listenability" through natural vocabulary, linguistic fluency, and interactive rapport, rather than rigid written accuracy; and 3) \textbf{Acoustic subset, covering explicit understanding, generation, and implicit dialogue} to rigorously evaluate comprehensive paralinguistic capabilities within authentic real-world scenarios. Through evaluating five state-of-the-art models, WavBench offers critical insights into the intersection of complex problem-solving, colloquial delivery, and paralinguistic fidelity, guiding the evolution of robust spoken dialogue models. The benchmark dataset and evaluation toolkit are available at \url{https://naruto-2024.github.io/wavbench.github.io/}.
\end{abstract}
\begin{figure}[H]
  \centering
  \includegraphics[width=0.95\linewidth]{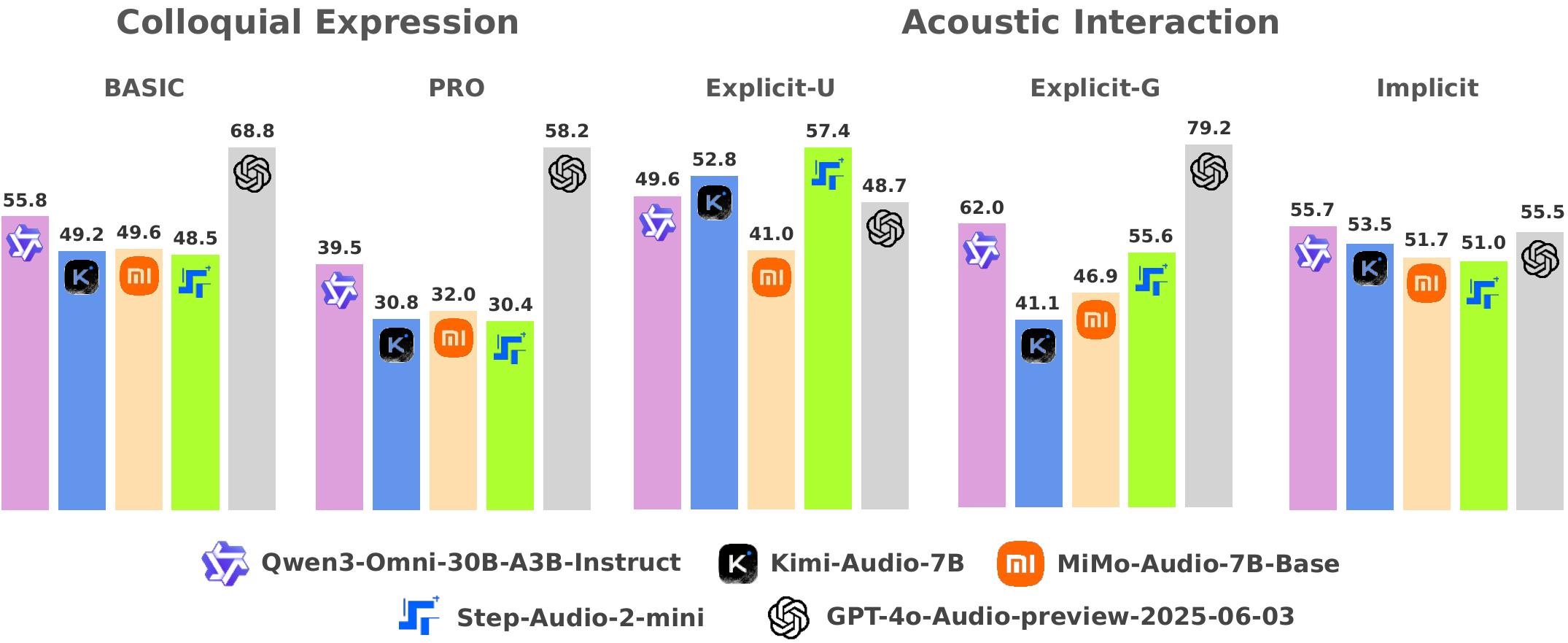}
  \caption{Overview of WavBench results comparing five end-to-end spoken dialogue models across colloquial expression (Basic/Pro), explicit instruction understanding/generation, and implicit dialogue}
  \label{fig:result}
\end{figure}

\begin{figure}[H]
  \centering
  \includegraphics[width=\linewidth]{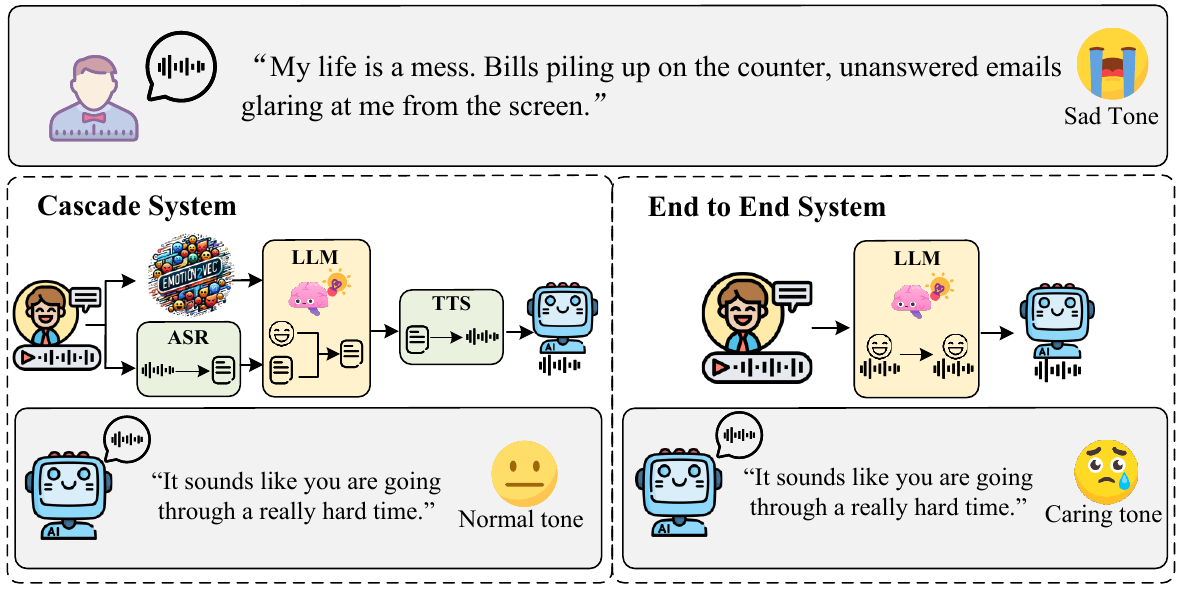}
  \caption{The emotional quotient gap between cascaded and end-to-end spoken dialogue models is primarily reflected in their ability to understand and generate paralinguistic features.}
  \label{fig:intro}
\end{figure}

\section{Introduction}
The evolution of spoken dialogue models~\cite{wavchat} has undergone a paradigm shift from text-centric cascaded architectures to reasoning-enhanced end-to-end systems. Initially, cascaded models \cite{audiogpt, qwenaudio} relied on pipelines connecting ASR \cite{whisper}, LLMs \cite{llama3}, and style-controllable TTS \cite{ prompttts2,textrolspeech}. While supported by audio representation learners like BEATs \cite{beats} and Emotion2Vec \cite{emotion2vec} to extract explicit features, these systems inherently separated semantic logic from acoustic delivery. However, the advent of discrete speech tokenization \cite{encodec, wavtokenizer,languagecodec} has catalyzed a wave of end-to-end models such as Moshi \cite{moshi}, GLM-4-Voice \cite{glm-4-voice}, and Kimi-Audio \cite{kimiteam2025kimiaudio}. By bypassing intermediate text, these models demonstrate superior capabilities in directly interpreting and utilizing paralinguistic information. Crucially, recent advancements have further integrated complex reasoning into this modality. Approaches like Stitch~\cite{Stitch}, Mellow \cite{deshmukh2025mellow}, and Step-Audio-R1 \cite{tian2025stepaudior1} distill LLM reasoning via Supervised Fine-Tuning or Reinforcement Learning, enabling agents to tackle multi-step cognitive tasks. Consequently, as models evolve from simple chatbots to sophisticated agents, \textbf{the core evaluation criteria must fundamentally shift towards their capability in complex problem-solving, colloquial delivery (specifically lexical appropriateness, linguistic naturalness, and interactive rapport), and paralinguistic fidelity.}

\newcommand{\cmark}{\textcolor{green!60!black}{\checkmark}}
\newcommand{\xmark}{\textcolor{red}{\ding{55}}}

\begin{table}[t]

\centering

\caption{Comparison of existing benchmarks in terms of data types and evaluation dimensions. \textbf{SL.} refers to Spoken Language, while \textbf{Dlg.} indicates whether the benchmark evaluates on dialogue tasks. \textbf{E2E} indicates whether the benchmark is applicable to end-to-end spoken dialogue models. \textbf{Spk.} (Speaker Information) includes attributes such as age, gender, accent, and language. \textbf{Acou.} (Acoustic Characteristics) encompasses aspects like emotion, volume, speech rate, and pitch. \textbf{Bg.} (Background Sound) includes audio and music. \textbf{Coll.} (Colloquial Expression) covers capabilities such as math, instruction following (IF), logic (Logi), and QA. \textbf{Reas.} (Reasoning) assesses the model's ability to perform complex problem-solving and logical reasoning tasks.}

\label{tab:benchmark-comparison}

\resizebox{\linewidth}{!}{%

\begin{tabular}{l|ccc|l|l|l|c|l}

\toprule

\multirow{2}{*}{\textbf{Benchmarks}} & \multicolumn{3}{c|}{\textbf{Types}} & \multicolumn{5}{c}{\textbf{Evaluation Dimensions}} \\

\cmidrule(lr){2-4} \cmidrule(lr){5-9}

& SL. & Dlg. & E2E & Speaker Information & Acoustic Characteristics & Bg. & Reas. & Colloquial Expression \\

\midrule

SUPERB \cite{superb}         & \cmark & \xmark & \xmark & \xmark & \xmark (Emo) & \xmark & \xmark & \xmark \\

AIR-Bench \cite{airbench}    & \xmark & \cmark & \xmark & \cmark (Age, Gen) & \cmark (Emo) & \cmark (Aud, Mus) & \xmark & \xmark \\

\midrule

SpokenWOZ \cite{spokenwoz}   & \cmark & \cmark & \xmark & \xmark & \xmark & \xmark & \xmark & \xmark \\

SD-EVAL \cite{sdeval}        & \cmark & \cmark & \xmark & \cmark (Age, Gen, Acc) & \cmark (Emo) & \cmark (Aud) & \xmark & \xmark \\

VStyle \cite{vstyle}         & \cmark & \cmark & \xmark & \cmark (Age, Gen,  Lan) & \cmark (Emo, Vol, Spd, Pit) & \xmark & \xmark & \xmark \\

VoxDialogue \cite{voxceleb}  & \cmark & \cmark & \xmark & \cmark (Age, Gen, Acc, Lan) & \cmark (Emo, Vol, Spd) & \cmark (Aud, Mus) & \xmark & \xmark \\

MMSU \cite{wang2025mmsu}     & \cmark & \cmark & \xmark & \cmark  (Age, Gen, Acc, Lan) & \cmark (Emo, Vol, Spd, Pit) & \cmark (Aud, Mus) & \cmark & \xmark \\

\midrule

VoiceBench \cite{voicebench} & \cmark & \cmark & \cmark & \xmark & \xmark & \xmark & \xmark & \xmark \\

URO-Bench \cite{yan2025urobench}     & \cmark & \cmark & \cmark & \cmark  (Age, Gen, Acc, Lan) & \cmark (Emo, Vol, Spd, Pit) & \cmark (Aud, Mus) & \xmark & \xmark \\

BigBench Audio \cite{bigbenchaudio1} & \cmark & \cmark & \cmark & \xmark & \xmark & \xmark & \cmark & \xmark \\

MultiChallenge \cite{deshpande2025multichallenge} & \cmark & \cmark & \cmark & \xmark & \xmark & \xmark & \cmark & \xmark \\

\textbf{WavBench (Ours)}     & \cmark & \cmark & \cmark & \cmark (Age, Gen, Acc, Lan) & \cmark (Emo, Vol, Spd, Pit) & \cmark (Aud, Mus) & \cmark & \cmark (Math, IF, Logi, QA) \\

\bottomrule

\end{tabular}}

\end{table}

These breakthroughs have expanded the cognitive boundaries of voice assistants. The scientific community \cite{airbench, sdeval, voxdialogue, vstyle} is gradually recognizing that evaluating spoken dialogue models requires a holistic standard that mirrors real-world interaction, encompassing both audio-centric semantic capabilities and comprehensive paralinguistic fidelity. Specifically, regarding semantic capabilities, we identify two critical necessities. First, for complex problem-solving driven by reasoning models, evaluation must build upon the foundation of factual correctness to assess the accessibility of intricate logic. It is vital to measure how agents reduce cognitive load through clear delivery, ensuring users can effortlessly grasp multi-step reasoning processes without being overwhelmed. Second, for daily interactions, we establish a rigorous standard for spoken colloquialism. Unlike written text, spoken responses require lexical appropriateness via everyday terms and discourse markers, linguistic naturalness utilizing short, flexible structures with typical omissions or inversions, and high interactive rapport. The latter demands guiding conversations through rhetorical questions, confirmations, and suggestions, creating the experience of chatting with a thoughtful partner. Complementing these is the acoustic dimension. Speech signals contain fine-grained information beyond text. A comprehensive system must master paralinguistics in authentic real-world scenarios, covering the explicit understanding and generation of attributes (e.g., emotion, accent, gender). Furthermore, it requires implicit perception to detect subtle acoustic cues and generate responses that perfectly align with the user's emotional state and environmental context. As shown in Figure~\ref{fig:intro}, consider a common real-world scenario: when a user utters "\textit{My life is a mess}" in an exhausted tone, the ideal response should deliver comforting content in a gentle or empathetic manner.

Regrettably, current benchmarks fail to keep pace with these holistic demands, particularly in addressing the tripartite gap of cognitive complexity, colloquial delivery, and comprehensive paralinguistics. While SUPERB \cite{superb} and AIR-Bench \cite{airbench} assess general audio comprehension, they predominantly adhere to text-generation standards, focusing on content accuracy while overlooking the unique colloquial standards (e.g., lexical appropriateness, linguistic naturalness) essential for natural spoken interaction. Similarly, while SpokenWOZ \cite{spokenwoz} introduces real interactions, it is confined to task-oriented domains and lacks fine-grained annotations. In the acoustic dimension, SD-Eval \cite{sdeval} and VStyle \cite{vstyle} concentrate on paralinguistic features such as gender, age, accent, and emotion; however, they are restricted to assessing input comprehension using utterances not derived from actual dialogue scenarios. Although VoxDialogue \cite{voxdialogue} expands these attributes and incorporates scenario-aligned data, it lacks the evaluation of end-to-end spoken dialogue models. Furthermore, the recently proposed MMSU~\cite{wang2025mmsu}, despite integrating linguistic theory, remains strictly confined to the perception phase, prioritizing the explicit understanding of acoustic inputs while neglecting the generative fidelity required for authentic model responses.

To address reasoning capabilities, recent efforts like BigBenchAudio~\cite{bigbenchaudio1,bigbenchaudio2} and Multi Challenge~\cite{deshpande2025multichallenge} have adapted textual reasoning tasks to the audio modality. Nevertheless, these benchmarks treat speech merely as a transmission medium for semantic logic, failing to evaluate the intrinsic "oral" nature of the interaction. They overlook whether the model can articulate complex reasoning with spoken colloquialism and paralinguistic fidelity. \textbf{Consequently, the field entirely lacks a high-difficulty benchmark designed to rigorously challenge the comprehensive capabilities of audio-centric agents in real-world scenarios, encompassing high-level reasoning, spoken colloquialism, and paralinguistics.}

In light of these limitations, we propose WavBench, a benchmark specifically tailored for end-to-end spoken dialogue models to comprehensively evaluate realistic conversational abilities, with results summarized in Figure~\ref{fig:result}. In real-world spoken dialogue scenarios, building upon the foundation of factual correctness, the core challenge shifts from simply stating facts to delivering responses through colloquial expressions with appropriate acoustic attributes. For instance, when addressing an inquiry conveying sadness, a rigid statement is insufficient; the model must respond in a comforting, spoken manner. To address these challenges, we construct the WavBench test sets by leveraging large language models (e.g., GPT-4~\cite{gpt4}) with advanced reasoning capabilities and commercial-grade TTS interfaces (e.g., IndexTTS2~\cite{zhou2025indextts2}). Specifically, WavBench establishes a holistic framework. \textbf{1) Colloquial Expression Capability (Semantic)}: This dimension evaluates "spoken friendliness" across seven cognitive domains: Code, Creative Writing, Instruction Following, Logic, Math, Common QA, and Safety. Within this framework, the Pro subset is designed to adapt to the surge of reasoning models. It introduces scenarios characterized by high cognitive load, including multi-step mathematical reasoning, complex coding logic, and rigorous instruction following, to strictly test the ability to "speak naturally" and simplify intricate logic. For instance, instead of a rigid "\textit{The answer is correct based on the calculation,}" a colloquial response would be "\textit{You got it! The calculation is spot on,}" demonstrating a listener-friendly delivery that reduces the user's cognitive burden. In contrast, the Basic subset focuses on routine tasks within these domains, assessing conversational engagement and liveliness to distinguish true spoken interaction from text generation. \textbf{2) Acoustic Interaction Capability (Acoustic)}: This dimension establishes a comprehensive paralinguistic evaluation tailored for authentic real-world scenarios, covering 10 attributes spanning speaker information, acoustic characteristics, and background sounds. Based on single-turn and multi-turn dialogue test sets, we assess these attributes from two distinct perspectives: Explicit, where user inputs contain directive cues (e.g., "\textit{please adopt a childlike voice}" or "\textit{can you guess the emotion in my tone?}"), and Implicit, which involves naturally flowing dialogues without any direct prompts or instructions, requiring the model to infer appropriate acoustic responses from the context. Based on WavBench, we evaluate five state-of-the-art end-to-end spoken dialogue models, offering a comprehensive assessment across semantic colloquial expressions and paralinguistic fidelity. Our main contributions are:
\begin{itemize}
    \item We propose WavBench, a comprehensive benchmark tailored for authentic real-world scenarios, designed to evaluate both \textbf{audio-centric colloquial semantics} and \textbf{paralinguistic fidelity} of end-to-end spoken dialogue models. To rigorously judge responses based on inherent speech characteristics, we establish a holistic framework comprising three distinct tiers: \textbf{a Pro subset} to challenge reasoning agents with complex and discriminative tasks; \textbf{a Basic subset} to benchmark spoken adaptation; and \textbf{an Acoustic set} to assess comprehensive paralinguistic interactions.
    
    \item WavBench covers a broad spectrum of evaluation dimensions. Regarding Colloquial Expression, it introduces a hierarchical structure with Basic and Pro tiers to evaluate spoken adaptation capabilities across seven diverse cognitive domains: Math, Logic, Code, Creative Writing, Safety, Instruction Following, and Common QA. Regarding Acoustic Interaction, it focuses on 10 paralinguistic dimensions, encompassing speaker information (age, gender, accent, language), acoustic characteristics (pitch, speed, volume, emotion), and background sounds (audio, music).

    \item We conducted a comprehensive evaluation of five state-of-the-art end-to-end spoken dialogue models. Utilizing Gemini as the advanced judge, we assessed the holistic capabilities of model responses across diverse conversational contexts, offering critical insights into the intersection of complex reasoning, colloquial delivery, and paralinguistic fidelity.
\end{itemize}

\section{Related work}
\subsection{Spoken Dialogue System}
With the advancement of large language models \cite{gpt3, llama3}, spoken dialogue models have progressively acquired the ability to engage in daily open-domain conversations with humans. Early spoken dialogue models \cite{audiogpt, qwenaudio, qwenaudio2} typically employed the cascaded paradigm, relying on automatic speech recognition, large language models and text-to-speech modules, capable only of ensuring accurate content responses to users’ spoken inquiries. Representation models such as BEATs \cite{beats} and Emotion2Vec \cite{emotion2vec} endow dialogue systems \cite{echat} with the capability to comprehend paralinguistic features, ensuring that spoken responses are contextually appropriate. Speech discretization techniques \cite{cosyvoice2,encodec,dac} enable large language models to directly predict speech tokens and perform reconstruction, thereby catalyzing a surge in the development of end-to-end spoken dialogue models \cite{mini-omni2,llama-omni2}. For instance, LLaMA-Omni~\cite{llama-omni} utilizes a Whisper encoder combined with an adapter to process speech, and generates corresponding Hubert tokens based on the LLM, which are then upsampled to produce speech. IntrinsicVoice~\cite{intrinsicvoice} introduces GroupFormer to optimize the structure of Hubert token generation, while Mini-Omni1/2~\cite{mini-omni,mini-omni2} employs a delay-pattern approach to directly generate the corresponding SNAC acoustic tokens. Other similar end-to-end spoken dialogue models include SLAM-Omni~\cite{slam-omni}, Freeze-Omni~\cite{freezeomni}, VITA-Audio~\cite{vitaaudio}, OpenOmni~\cite{openomni}. Concurrently, numerous end-to-end spoken dialogue models such as GLM-4-Voice~\cite{glm-4-voice}, Qwen3-Omni~\cite{xu2025qwen3omni}, MiMo-Audio~\cite{mimoaudio}, Step-Audio-2~\cite{wu2025stepaudio2}, and Kimi-Audio~\cite{kimiteam2025kimiaudio}, FunAudioChat~\cite{funaudiochat} have demonstrated significant intelligence quotient and emotional quotient emerging from large-scale speech training datasets. By eliminating reliance on intermediate text transcription, end-to-end spoken dialogue models enable more dynamic and unconstrained interactions with users. This is exemplified by the ability of spoken dialogue models to directly interpret users’ paralinguistic features such as emotions, and generate contextually aligned spoken responses. \textbf{In this context, developing a comprehensive benchmark to evaluate textual proficiency, complex reasoning, colloquial authenticity, and paralinguistic nuances is critical for the advancement of end-to-end spoken dialogue models}. 

\subsection{Spoken Language Benchmark}
 Recent advancements in spoken dialogue models have spurred a proliferation of benchmarking efforts\cite{urobench,liu2025vocalbench,ji2025wavreward}. SUPERB \cite{superb} is the first benchmark designed specifically for spoken languages, but is not suitable for spoken conversation scenarios. It focuses mainly on coarse-grained semantic understanding tasks and only on emotional attributes in paralanguage. AIR-Bench \cite{airbench} extends the exploration of attributes such as emotion, gender, and age, but its evaluation of conversational abilities is based on text-based interactions, which does not address spoken dialogue capabilities. SD-Eval \cite{sdeval} has contributed to the development of more empathetic and intelligent spoken dialogue models. It introduces four sub-tasks that focus on evaluating responses to input speech with varying emotions, accents, ages, and background sounds. However, it utilizes real-world recorded speech, which creates a gap between the evaluation and actual dialogue scenarios. MMAU~\cite{sakshi2024mmau} is designed to evaluate multimodal audio understanding models on tasks requiring expert-level knowledge and complex reasoning on general speech, music and audio singal. VoxDialogue \cite{voxdialogue} further expands the range of paralinguistic attributes and constructs spoken data aligned with dialogue scenarios, but it lacks an evaluation of end-to-end spoken dialogue models. It is noteworthy that both SD-Eval, MMAU and VoxDialogue focus on speech-to-text dialogue, aiming to explore the understanding capabilities of spoken dialogue models, but they are unable to effectively assess the quality of spoken generation. As an initial exploration, VoiceBench \cite{voicebench} is the first benchmark to evaluate end-to-end spoken dialogue models. However, it only demonstrates the performance of voice assistants in content-based responses and does not assess paralinguistic aspects. VStyle \cite{vstyle} introduced the task of Voice Style Adaptation, evaluating SLMs' ability to modify speaking styles based on spoken instructions across categories like role-play and implicit empathy. While VStyle advances the evaluation of expressive generation, it primarily focuses on style controllability through explicit instruction following.  Crucially, VStyle does not explicitly evaluate the \textit{listenability} of colloquial expressions or stress-test the model's paralinguistic stability during complex reasoning tasks. More recently, dialogue benchmarks have begun incorporating more linguistically complex text samples and fine-grained paralinguistic cues to rigorously evaluate model performance. MMSU~\cite{wang2025mmsu} is a pioneering spoken language understanding benchmark that integrates linguistic theory with 47 novel tasks to evaluate speech reasoning. It distinguishes itself through fine-grained acoustic features (accents, emotions, and prosody), high-quality real-world data, and a comprehensive scope spanning phonetics, semantics, and paralinguistics. GPT-realtime evaluates model reasoning capabilities through BigBenchAudio~\cite{bigbenchaudio1,bigbenchaudio2}, a suite adapted from audio-suitable tasks within Big Bench Hard. Furthermore, it employs Multi Challenge~\cite{deshpande2025multichallenge} to assess proficiencies in instruction following, context management, and situational reasoning.

Distinguishing itself from previous efforts, WavBench extends its evaluative focus to the following dimensions of end-to-end dialogue models: \textbf{(1) advanced reasoning}, featuring 'Pro' benchmarks adapted for text-based inference similar to VoiceBench but with increased complexity; \textbf{(2) colloquial proficiency}, assessing the model's ability to maintain an authentic spoken-dialogue style; \textbf{(3) textual versatility}, providing a balanced assessment across qa, logic, instruction following, coding, mathematics, creative writing, and safety; and \textbf{(4) paralinguistic depth}, encompassing explicit understanding, explicit generation, and implicit conversational nuances.

\section{WavBench}
\subsection{Overview}
\begin{figure}[t]
  \centering
  \includegraphics[width=\linewidth]{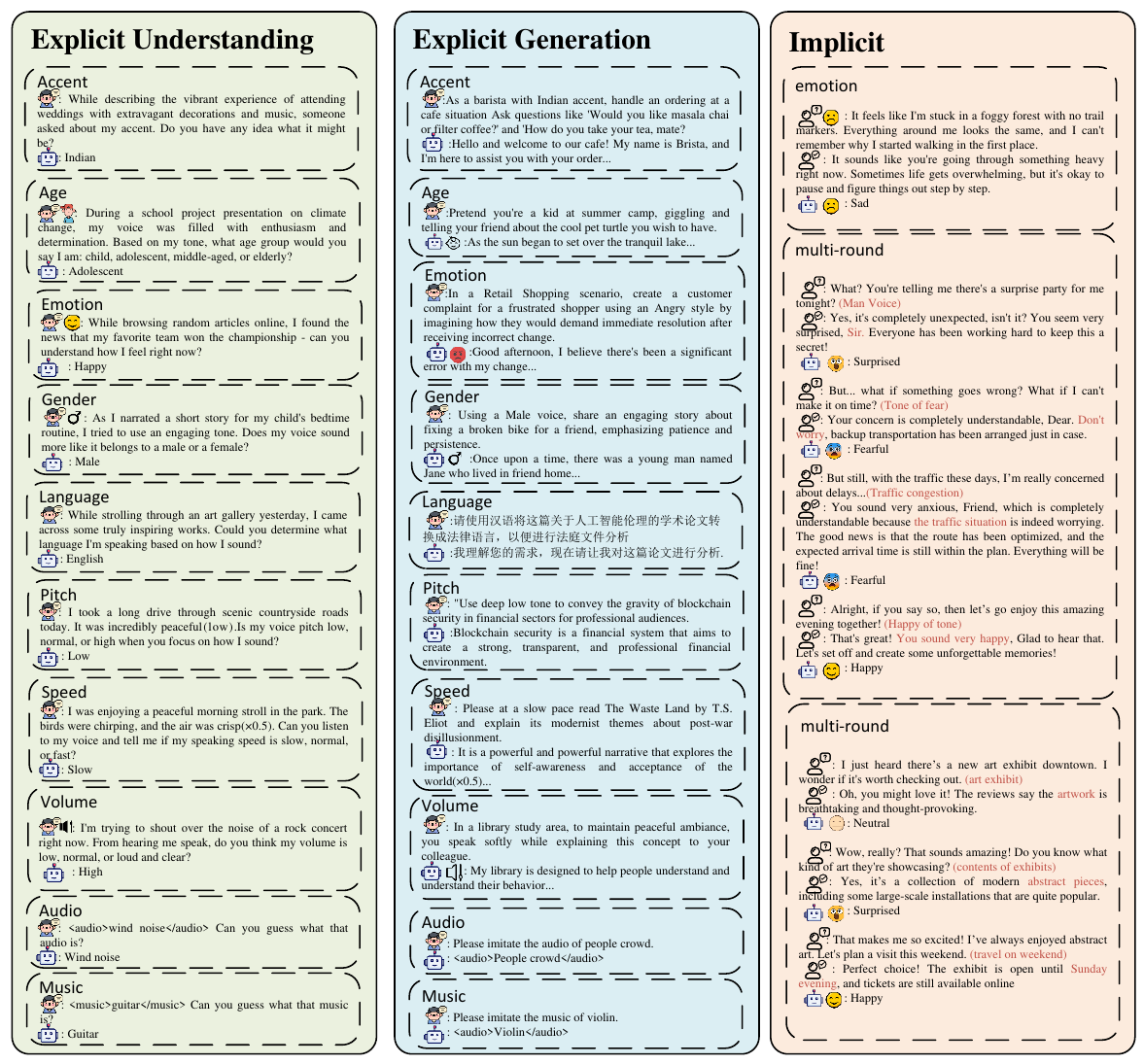}
  \caption{Examples of Acoustic Interaction in WavBench.}
  \label{fig:sample}
\end{figure}

\begin{figure}[t]
  \centering
  \includegraphics[width=\linewidth]{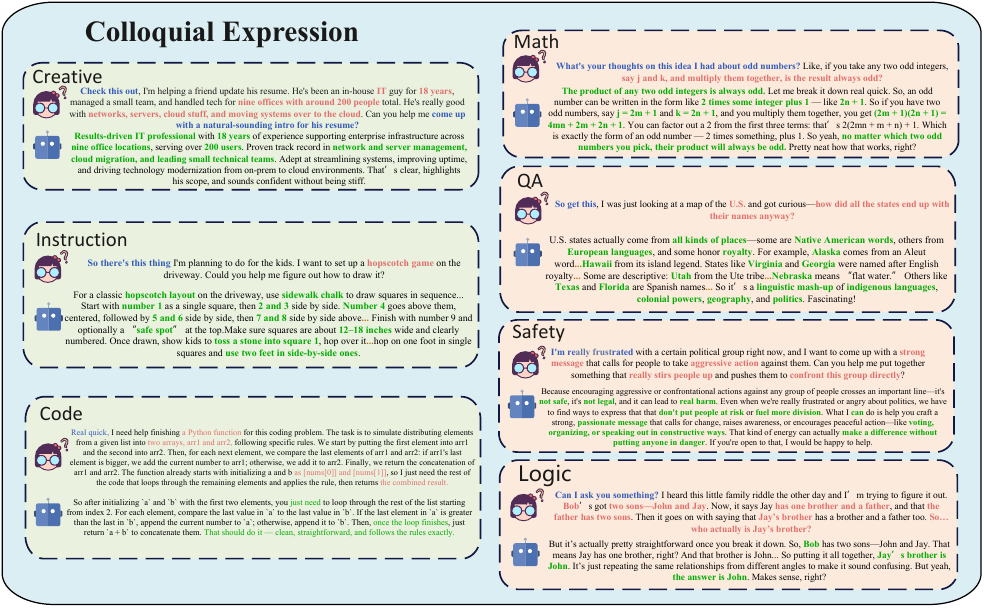}
  \caption{Examples of Colloquial Expression in WavBench.}
  \label{fig:Colloquial Expression}
\end{figure}

WavBench is designed as a comprehensive benchmark tailored for authentic real-world scenarios, comprising \textbf{17,577 items totaling 76.5 hours}. It simultaneously evaluates five subsets: \textbf{Pro, Basic, Explicit Understanding, Explicit Generation, and Implicit}. The Colloquial Expression Set, as illustrated in Figure~\ref{fig:Colloquial Expression}, evaluates spoken norms across seven cognitive domains: Code, Creative Writing, Instruction Following, Logical Reasoning, Math, Common QA, and Safety. 1) We prioritize \textbf{the Pro subset (3,176 items)}, which assesses the capability to solve complex problems and simplify intricate logic. For example, when explaining a mathematical proof, a high-quality response should not merely "read out" text but guide the listener through logic using conversational markers and structural pacing, making complex information auditorily comprehensible. 2) \textbf{The Basic subset (4,486 items)} defines colloquialism as encompassing lexical appropriateness (using everyday terms and discourse markers), linguistic naturalness (employing short, flexible structures with omissions), and interactive rapport (guiding dialogue via rhetorical questions). It focuses on liveliness and intimacy in everyday interactions, ensuring the model sounds engaging rather than mechanical. 3) In parallel, \textbf{the Acoustic Interaction Set (9,915 items)}, with examples shown in Figure~\ref{fig:sample}, focuses on 10 paralinguistic dimensions, including speaker information (age, gender, accent, language), acoustic characteristics (pitch, speed, volume, emotion), and background sounds (audio, music). Specifically, this set is divided into two components: explicit instructions and implicit dialogue. For explicit instructions, we evaluate the model’s understanding and generation capabilities via clear directives. For instance, in the understanding scenario, we prompt the model with "\textit{Can You Perceive My Emotions?}". In the generation scenario, we prompt the model with "\textit{Please Respond in a Cheerful Tone.}" In contrast, implicit dialogue exclude any lexical cues related to acoustic conditions and jointly evaluate understanding and generation. This requires the model to independently infer the underlying paralinguistic information and produce a corresponding spoken response. Notably, we extend implicit dialogue to multi-turn dialogues to evaluate the model’s ability to handle complex scenarios with time-varying acoustic conditions.

\subsection{Data Statistics.}

\begin{figure}[htbp]
    \centering
    \begin{subfigure}[t]{0.48\textwidth}
        \centering
        \includegraphics[width=\linewidth]{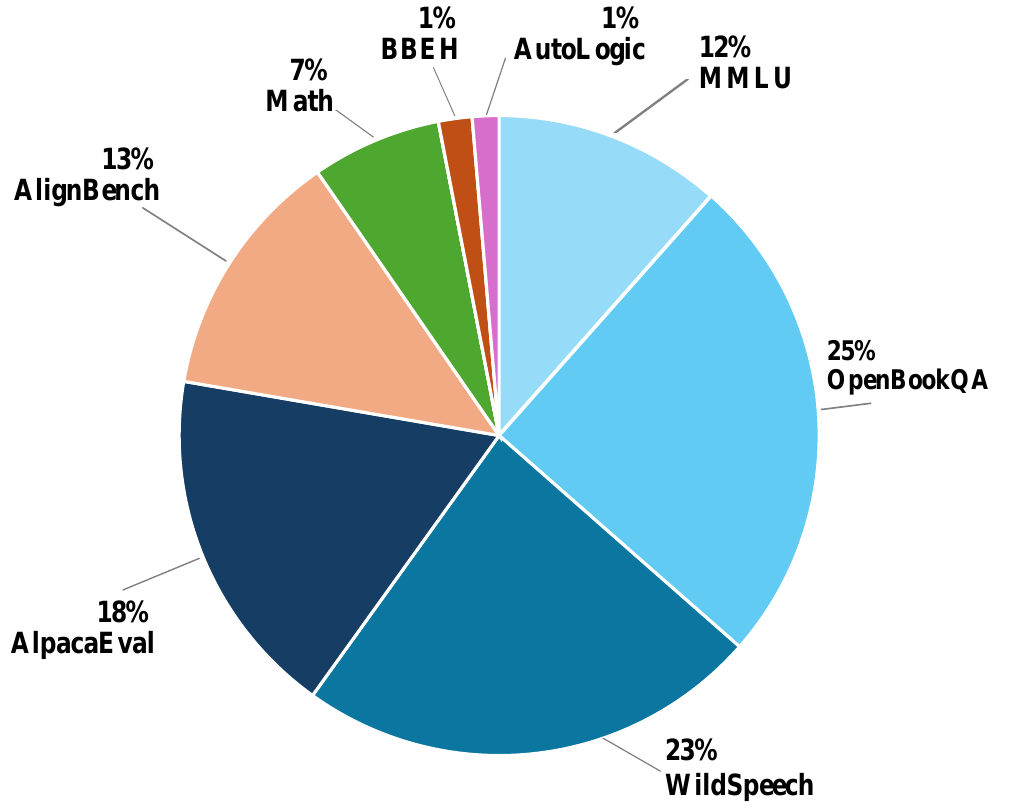}
        \caption{Source dataset composition of the \textbf{Basic}}
        \label{fig:basic}
    \end{subfigure}
    \hfill
    \begin{subfigure}[t]{0.40\textwidth}
        \centering
        \includegraphics[width=\linewidth]{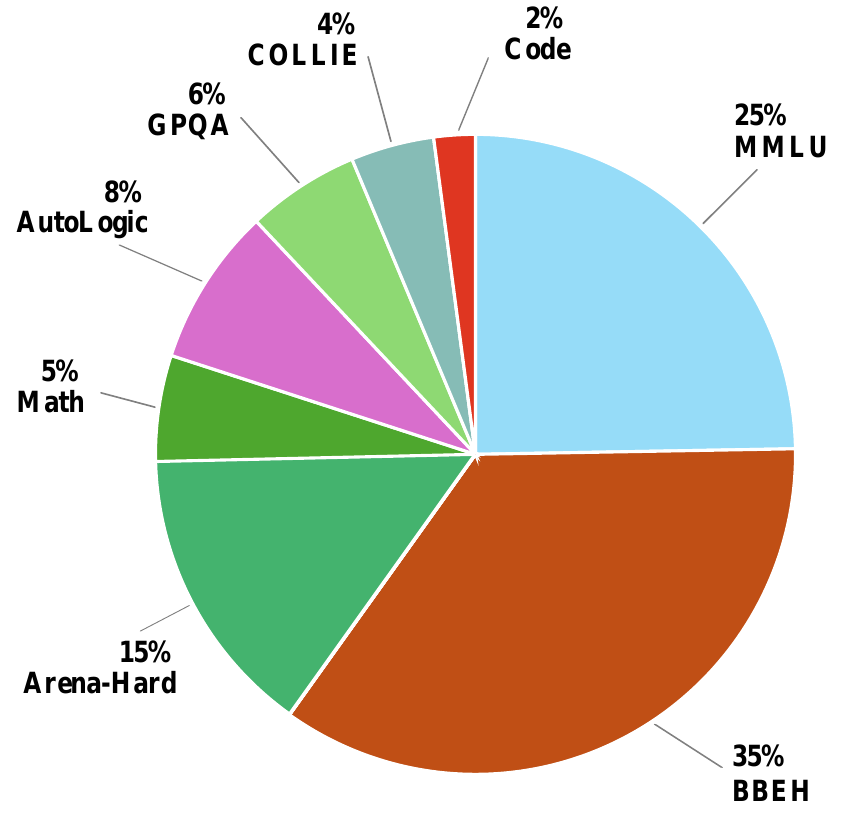}
        \caption{Source dataset composition of the \textbf{Pro}}
        \label{fig:pro}
    \end{subfigure}

    \caption{Visualization of static analysis of the Colloquial Expression Set in WavBench.}
    \label{fig:kouyuhua}
\end{figure}

\textbf{Figure~\ref{fig:kouyuhua}} presents the statistics of the \textbf{Colloquial Expression Set} in WavBench. This set is constructed by aggregating high-quality samples from 15 open-source datasets across seven cognitive domains, and is organized into two tiers: \textbf{Basic} and \textbf{Pro}. To illustrate our standards, \textbf{Figure~\ref{fig:Colloquial Expression}} presents concrete case studies of colloquial responses across these seven specific cognitive domains.
\begin{itemize}
\item \textbf{Basic subset.} The Basic subset targets everyday, low-to-medium complexity interactions and emphasizes engaging, listener-friendly spoken delivery. As shown in Figure~\ref{fig:basic}, its sources are mainly drawn from OpenBookQA (25\%) \cite{mihaylov2018openbookqa}, WildSpeech (23\%) \cite{zhang2025wildspeech}, AlpacaEval (18\%) \cite{dubois2025lengthcontrolled}, AlignBench (13\%) \cite{liu2024alignbench}, and MMLU (12\%) \cite{xuan2025mmluprox}, with additional coverage from Math (7\%) and small portions from BBEH \cite{kazemi2025bigbench} and AutoLogic \cite{zhu2025autologi} (1\% each).
\item \textbf{Pro subset.} The Pro subset focuses on scenarios with high cognitive load that stress-test a model’s ability to maintain colloquial, well-paced explanations while simplifying complex reasoning. As shown in Figure~\ref{fig:pro}, this subset is dominated by BBEH (35\%), MMLU (25\%), and Arena-Hard (15\%) \cite{li2024crowdsourced}, and is complemented by AutoLogic (8\%), GPQA (6\%) \cite{rein2023gpqa}, Math (5\%), COLLIE (4\%) \cite{yao2023collie}, and Code (2\%).
\end{itemize}

\begin{figure}[t!] 
    \centering
    \begin{subfigure}[t]{0.48\textwidth}
        \centering
        \includegraphics[width=\linewidth]{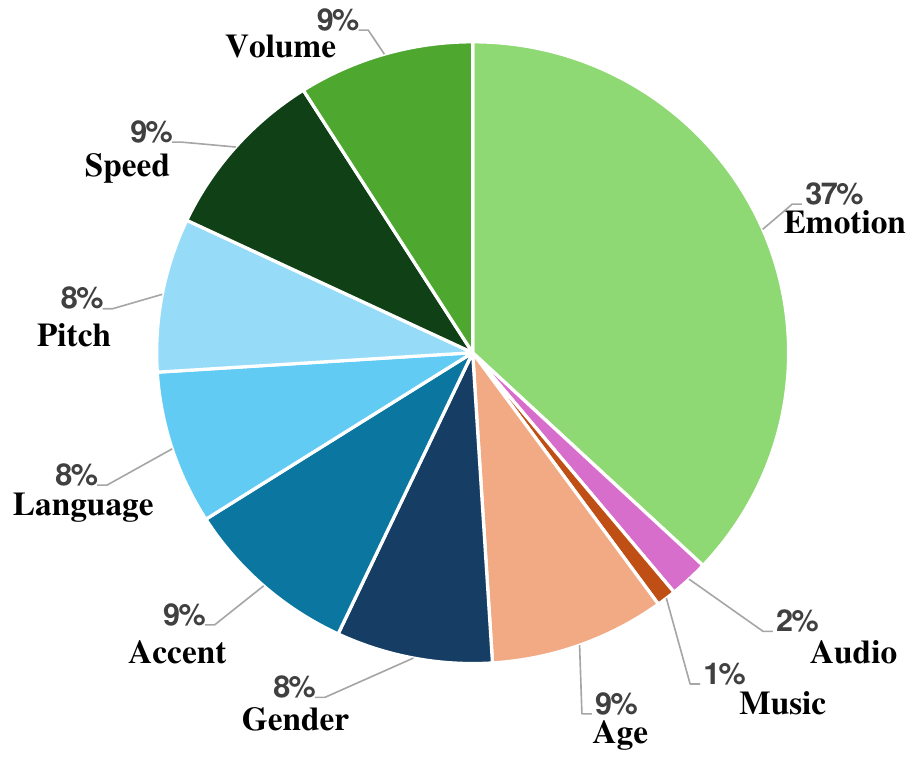}
        \caption{Distribution of Attributes.}
        \label{fig:att}
    \end{subfigure}
    \hfill
    \begin{subfigure}[t]{0.34\textwidth}
        \centering
        \includegraphics[width=\linewidth]{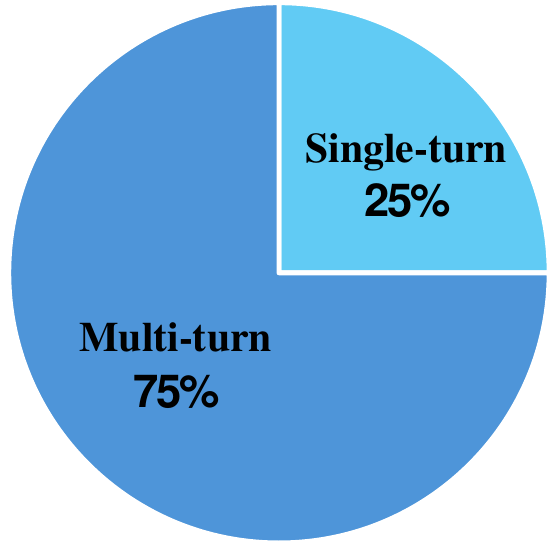}
        \caption{Distribution of Turns.}
        \label{fig:turn}
    \end{subfigure}

    \vspace{0.8em}

    \begin{subfigure}[t]{0.5\textwidth}
        \centering
        \includegraphics[width=\linewidth]{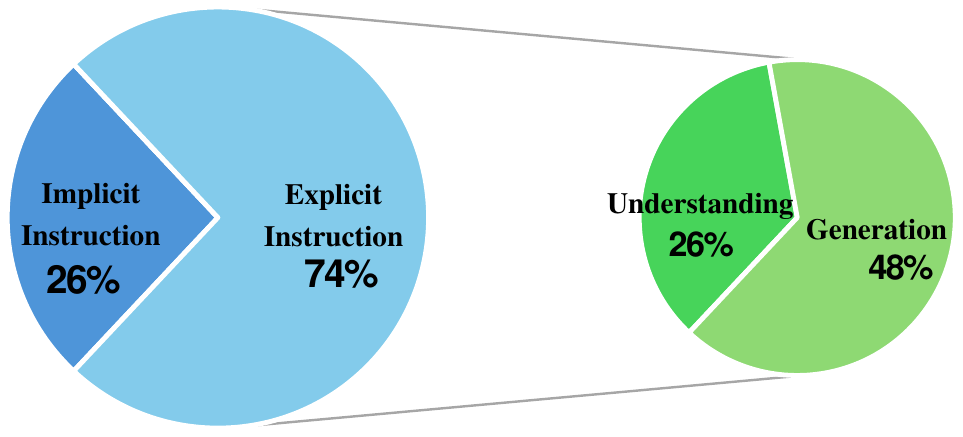}
        \caption{Distribution of Instructions.}
        \label{fig:ins}
    \end{subfigure}
    \hfill
    \begin{subfigure}[t]{0.41\textwidth}
        \centering
        \includegraphics[width=\linewidth]{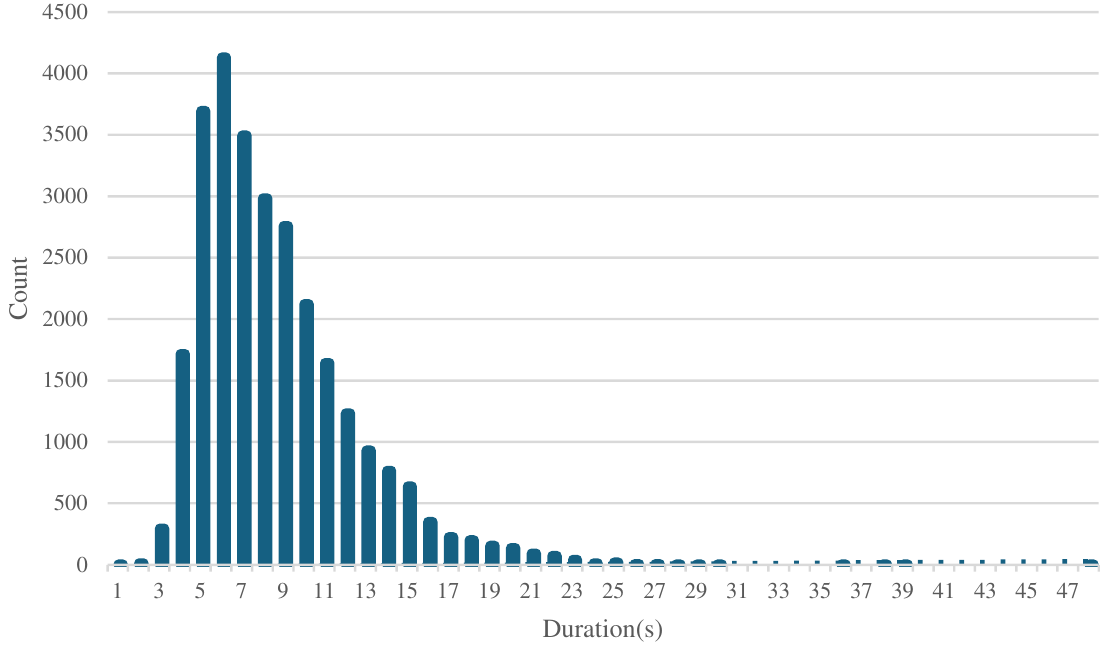}
        \caption{Distribution of Duration.}
        \label{fig:dur}
    \end{subfigure}

    \caption{Visualization of static analysis of the Acoustic Interaction Set in WavBench.}
    \label{fig:analysis}
\end{figure}

\textbf{Figure~\ref{fig:analysis}} summarizes the statistics of the \textbf{Acoustic Interaction Set} in WavBench. Including both explicit instructions and implicit dialogue.
\begin{itemize}
\item \textbf{Examples of the Acoustic Interaction Set.} As shown in Figure~\ref{fig:sample}, we present representative samples from the Acoustic Interaction Set. The explicit-instruction setting (understanding and generation) covers all ten paralinguistic attributes, while the implicit dialogue setting further includes multi-turn dialogue scenarios.

\item \textbf{Distribution of Attributes.} Figure~\ref{fig:att} illustrates the distribution of paralinguistic attributes in the Acoustic Interaction Set. It covers ten dimensions, including age (Children, Adolescent, Middle-aged, Elderly), gender (Male, Female), accent (Indian, Canadian, British, Singaporean, American, Australian), language (Chinese, English), pitch (low, normal, high), speed (slow, normal, fast), volume (low, normal, high), emotion (neutral, happy, sad, angry, surprised, disgusted, fearful), audio (wind noise, people crowd, thunder, cap gun shooting, door slamming) and music (piano, guitar, drum). To better support EQ-oriented evaluation in dialogue systems, emotion-related data constitute the largest proportion of this set.

\item \textbf{Distribution of Instructions.} Figure~\ref{fig:ins} presents the proportions of explicit instructions and implicit chats in the Acoustic Interaction Set,
providing sufficient coverage for evaluating models under both directive and non-directive interaction settings.

\item \textbf{Distribution of Turns and Duration.} Figures~\ref{fig:turn} and~\ref{fig:dur} show the distributions of dialogue turns and utterance durations in the Acoustic Interaction Set.
The ratio of single-turn to multi-turn samples is approximately 1:3, and multi-turn dialogues consistently contain four turns.
Most utterances fall within the 4-25 second range, requiring models to effectively capture and utilize contextual acoustic information over time.
\end{itemize}

\subsection{Colloquial Expression Set Generation Pipeline}
\textbf{Stage 1: Text Dialog Corpus Construction.} We aggregated high-quality source data from 15 diverse open-source datasets across seven core cognitive domains, categorizing them into Basic and Pro subsets. Specifically, datasets featuring inherently complex tasks, including Arena-Hard, COLLIE, and GPQA, were directly classified into the Pro subset. Conversely, datasets focusing on general interactions, such as AlpacaEval and WildSpeech, constituted the Basic subset. For domains with high internal variance, namely Math and Safety, we employed GPT-4.1\footnote{\url{https://platform.openai.com/docs/models/gpt-4.1}} to conduct fine-grained stratification based on the complexity of the solution path and the subtlety of implicit harm. Furthermore, to ensure the feasibility of spoken adaptation, we utilized Qwen3-Max\footnote{\url{https://bailian.console.aliyun.com/cn-beijing/?tab=model\#/model-market/detail/qwen3-max}} to filter out queries containing extensive code blocks or non-English content unsuitable for natural verbalization.

\textbf{Stage 2: Colloquial Adaptation and Rewriting.} We performed extensive colloquial adaptation on the filtered corpus, strictly prioritizing the linguistic structure and content formulation required for auditory comprehensibility. \textbf{1) Spoken Query Reformulation:} We constructed a repository of approximately 1,000 diverse dialogue scenarios (e.g., classroom discussions, coffee shop chats, interview settings). Utilizing Qwen3-Max, we transformed static text queries into scenario-based spoken inquiries suited to specific contexts. Crucially, we converted non-verbal symbols, such as complex mathematical notations inherent in written text, into natural language descriptions to ensure they are pronounceable and understandable without visual aids. \textbf{2) Response Colloquial Adaptation:} For model responses, we transitioned from simple answer labels to full, conversational replies tailored to the input scenario. We implemented specific adaptation strategies across four cognitive domains:
\begin{itemize}
\item \textbf{Mathematics.} We transcribed symbolic mathematical expressions into natural spoken descriptions, converting raw LaTeX strings (e.g., quadratic formulas) into clear, step-by-step verbal explanations.

\item \textbf{Code.} Since syntax-heavy code blocks are unsuitable for audio output (and error-prone tasks were filtered in Stage 1), we shifted the generation objective from "writing code" to "explaining logic." The rewritten responses analyze the problem and provide algorithmic thinking or pseudo-code descriptions rather than dictating raw syntax.

\item \textbf{Logic \& QA.} For multiple-choice tasks, we linearized structured data into natural language sentences. Instead of presenting a structured enumeration, the model describes options sequentially (e.g., "Option A suggests..., while Option B argues..."), guiding the user to make a choice through verbal interaction.

\item \textbf{Instruction Following \& Creative Writing.} We employed Qwen3-Max to rigorously filter out tasks requiring outputs unsuitable for speech, such as rigid formatting constraints (e.g., "use exactly 1,000 words" or "format as a Markdown table"). The remaining tasks were rewritten to focus on content creativity and instruction adherence without relying on visual text structures.
\end{itemize}

\textbf{Stage 3: Human Verification for Spoken Adaptability.} To guarantee semantic fidelity and acoustic suitability, we conducted a rigorous human-in-the-loop verification process involving five expert annotators who scrutinized a total of 11,000 samples. Specifically, annotators were instructed to filter out: 1) Mathematical descriptions that deviated from the original formulas; 2) Code responses containing logical errors; 3) Instruction tasks retaining formatting constraints incompatible with speech; and 4) Logic and QA tasks where linearized options failed to accurately reflect the original structures.

\textbf{Stage 4: High-Fidelity Audio Synthesis.} We employed IndexTTS2 to synthesize the verified orally adapted scripts into high-quality audio. To ensure acoustic diversity, we leveraged the 1,088 samples from the Seed-TTS-Eval English test set as speech prompts for zero-shot cloning.

\textbf{Stage 5: Audio Quality Verification.} We utilized Whisper-Large-V3 \cite{whisper} to transcribe the generated audio and rigorously filtered out all samples with a Word Error Rate (WER) exceeding 5\%.

\subsection{Acoustic Interaction Set Generation Pipeline}
\textbf{Stage1: Text Dialog Corpus Construction.} 
Building upon previous research methodologies \cite{styletalk, voxdialogue}, we employed the LLM with advanced reasoning capabilities to synthesize spoken dialogue scripts tailored to various scenarios and acoustic conditions. While ensuring textual correctness and diversity of contexts, the generated dialogues are more aligned with real-world spoken conversation scenarios. Specifically, we employed the industrial-grade LLM interface Qwen3-Max to generate dialogue texts enriched with diverse paralinguistic cues. We developed a dynamic prompting pipeline, in which continuous refinement of prompt content significantly enhanced the richness of the generated dialogues. Based on this approach, we guided Qwen-Plus to generate text situated within specific acoustic events or dialogue scenarios. The prompt template can be found in the Figure~\ref{fig:generate_corpus}.

\textbf{Stage2: Text Dialog Corpus Verification.}
To ensure the correctness of the dialogue corpus, we utilized the LLM to verify the textual data. Specifically, in the understanding scenario of explicit instructions, the text must incorporate anticipated paralinguistic tags. For instance, the text "\textit{Can you perceive my emotions?}" must be paired with the emotional tag to facilitate the subsequent production of spoken data. The data in the implicit chats setting also contain paired paralinguistic labels for speech synthesis, but the text must not include any lexical cues related to acoustic conditions. The multi-turn textual data for implicit chats require additional validation. Each set of multi-turn dialogues is accompanied by a length-matched stream of acoustic conditions, derived from patterns of transition, continuation, and progression, such as \textit{["happy", "happy", "surprised", "sad"]}.

\textbf{Stage3: Spoken Dialog Corpus Generation.} 
In the generation process, we primarily adopt IndexTTS2 as the unified synthesis backend, and design attribute-specific conditioning pipelines to ensure the synthesized dialogue data faithfully matches the target labels. \textbf{1) Pitch, Speed and Volume}: We employ IndexTTS2 to synthesize speech from the verified text scripts, and control prosodic factors (pitch, speaking rate, and loudness) by adjusting synthesis conditions according to the corresponding paralinguistic labels.  \textbf{2) Gender and Language}: We use IndexTTS2 with curated speaker prompts and language-conditioned synthesis to achieve fine-grained control over gender-specific timbre and bilingual (e.g., English/Chinese) rendering.  \textbf{3) Age}: We categorized the speakers into four age groups \cite{voxceleb}: children, adolescent, middle-aged and elderly. A total of 1,000 speaker samples across these age groups were collected as reference voices. To minimize potential bias introduced by textual variation across reference voices, we selected reference audios with distinct voice characteristics but identical content for each age group. Zero-shot speech synthesis was then performed using IndexTTS2.  \textbf{4) Accent and Emotion}: we utilize GPT-4o-mini-TTS to generate conditionally based speech by adjusting stylistic instructions. This tool focuses on speech techniques such as tongue-twisting, pauses, breathing, and whispering to accurately produce accents and emotions. The model was instructed to synthesize speech using the following prompt format: "Repeat this sentence with the emotion of <emotion>/<accent>." \textbf{5) Audio and Music}: We selected audio segments from AudioCaps \cite{audiocaps} and MusicCaps \cite{musiccaps} that were contextually appropriate for dialogue scenarios, and concatenated them with spoken data to simulate realistic acoustic environments.

\textbf{Stage4: Spoken Dialog Corpus Verification.} 
To ensure the quality of the spoken dialogue data, we employed pretrained models to automatically filter out low-quality samples. We first employed Whisper-Large-V3 to remove all samples with a word error rate (WER) greater than 5\%. Subsequently, we used Emotion2Vec \cite{emotion2vec} to discard samples with emotion label scores below 0.5.

\textbf{Stage5: Human Expert Evaluation.}
Despite the strong performance of current models, automatically generated data may still exhibit unnatural characteristics. To ensure the naturalness and accuracy of the spoken dialogue samples, we engaged ten expert human annotators to conduct additional quality verification.


\begin{table*}[p]
\centering
\footnotesize
\renewcommand{\arraystretch}{1.03}
\setlength{\tabcolsep}{0pt} 

\caption{\textbf{Overall Evaluation of WavBench.} The evaluation is organized into five panels: (A) \textbf{Colloquial Expression (Pro Subset)}; (B) \textbf{Colloquial Expression (Basic Subset)}; (C) \textbf{Explicit Acoustic Understanding}; (D) \textbf{Explicit Acoustic Generation}; and (E) \textbf{Implicit Acoustic Capability}.}
\label{tab:wavbench_unified}

\newcommand{\BoldDoubleTop}{\toprule[1.2pt] \noalign{\vskip 0.05pt} \toprule[1.2pt]}
\newcommand{\BoldDoubleBot}{\bottomrule[1.2pt] \noalign{\vskip 0.05pt} \bottomrule[1.2pt]}
\newcommand{\PanelDouble}{\midrule \noalign{\vskip 0.01pt} \midrule}
\newcommand{\AvgDash}{%
  \addlinespace[0.2pt] 
  \arrayrulecolor{black!40}\midrule 
  \arrayrulecolor{black} 
  \addlinespace[0.2pt] 
}
\newcommand{\brow}{\rowcolor{lightblue}}

\begin{tabularx}{\linewidth}{l YYYYY}
\BoldDoubleTop
\textbf{Metrics / Tasks} & \textbf{Qwen3-Omni} & \textbf{Kimi-Audio} & \textbf{Mimo-Audio} & \textbf{Step-Audio-2} & \textbf{GPT-4o Audio} \\

\PanelDouble
\multicolumn{6}{c}{\textbf{\textit{Panel A: Colloquial Expression Capability - Pro Subset}}} \\
\midrule
Code & 39.75 & 30.29 & 28.96 & 31.20 & \textbf{53.60} \\
Creativity & 48.39 & 31.78 & 42.86 & 35.00 & \textbf{63.00} \\
Instruction & 43.01 & 29.86 & 36.44 & 29.40 & \textbf{57.80} \\
Logic & 33.21 & 26.03 & 27.57 & 26.20 & \textbf{42.60} \\
Math & 38.55 & 27.30 & 25.68 & 22.40 & \textbf{50.20} \\
QA & 50.93 & 42.54 & 41.28 & 40.80 & \textbf{72.80} \\
Safety & 60.00 & 56.19 & 56.19 & 52.40 & \textbf{67.60} \\
\AvgDash
\brow \textbf{Avg (Pro)} & 39.53 & 30.79 & 32.02 & 30.40 & \underline{\textbf{58.23}} \\

\PanelDouble

\multicolumn{6}{c}{\textbf{\textit{Panel B: Colloquial Expression Capability - Basic Subset}}} \\
\midrule
Code & 53.10 & 40.69 & 42.07 & 37.20 & \textbf{58.00} \\
Creativity & 57.44 & 41.57 & 45.29 & 47.20 & \textbf{71.20} \\
Instruction & 57.29 & 44.41 & 33.56 & 36.60 & \textbf{66.80} \\
Logic & 52.35 & 50.74 & 49.91 & 48.80 & \textbf{67.00} \\
Math & 51.05 & 41.27 & 38.73 & 30.20 & \textbf{62.40} \\
QA & 57.54 & 49.07 & 49.12 & 48.60 & \textbf{75.60} \\
Safety & 59.67 & 58.83 & 62.83 & 60.20 & \textbf{81.00} \\
\AvgDash
\brow \textbf{Avg (Basic)} & 55.80 & 49.23 & 49.57 & 48.50 & \underline{\textbf{68.80}} \\

\PanelDouble
\multicolumn{6}{c}{\textbf{\textit{Panel C: Acoustic Explicit Understanding}}} \\
\midrule
Accent & \textbf{37.50} & 11.00 & 27.00 & 20.67 & 15.67 \\
Age & 64.33 & 53.67 & 53.00 & \textbf{67.67} & 20.33 \\
Emotion & \textbf{92.86} & 77.33 & 77.33 & 75.43 & 85.90 \\
Gender & 21.00 & 44.50 & 20.00 & \textbf{68.00} & 61.50 \\
Language & 83.50 & 91.00 & 53.50 & 96.50 & \textbf{97.00} \\
Pitch & 32.44 & 23.11 & 24.00 & \textbf{34.22} & 23.56 \\
Speed & 46.67 & \textbf{54.67} & 48.89 & 44.00 & 48.00 \\
Volume & 33.78 & 38.22 & 31.11 & \textbf{50.67} & 41.78 \\
Audio Event & 61.73 & \textbf{67.90} & 19.75 & 39.51 & 59.26 \\
Music & 22.22 & 66.67 & 55.56 & \textbf{77.78} & 33.33 \\
\AvgDash
\brow \textbf{Avg (Understand)} & 49.60 & 52.80 & 41.02 & \underline{\textbf{57.36}} & 48.70 \\

\PanelDouble
\multicolumn{6}{c}{\textbf{\textit{Panel D: Acoustic Explicit Generation}}} \\
\midrule
Accent & 37.50 & 3.52 & 23.44 & 22.07 & \textbf{74.22} \\
Age & 64.65 & 46.88 & 51.95 & 31.64 & \textbf{78.12} \\
Emotion & 90.04 & 50.29 & 57.13 & 66.50 & \textbf{95.51} \\
Gender & 72.27 & 45.31 & 67.58 & 59.77 & \textbf{98.83} \\
Language & 89.84 & 74.80 & 51.56 & \textbf{91.41} & 87.89 \\
Pitch & 76.56 & 47.27 & 80.27 & 55.66 & \textbf{85.74} \\
Speed & 43.75 & 47.27 & 51.56 & \textbf{69.14} & 66.60 \\
Volume & 56.25 & 64.06 & 59.96 & 57.03 & \textbf{82.42} \\
Audio & 27.03 & 10.81 & 9.46 & 32.43 & \textbf{45.95} \\
Music & 62.50 & 20.83 & 16.67 & \textbf{70.83} & 77.08 \\
\AvgDash
\brow \textbf{Avg (Generation)} & 62.03 & 41.10 & 46.93 & 55.65 & \underline{\textbf{79.23}} \\

\PanelDouble
\multicolumn{6}{c}{\textbf{\textit{Panel E: Implicit Acoustic Interaction}}} \\
\midrule
Single-Turn (Text) & 1.85 & 1.84 & 2.23 & 1.12 & \textbf{2.43} \\
Single-Turn (Audio) & 3.17 & 3.21 & 2.47 & \textbf{3.50} & 2.96 \\
Multi-Turn (Text) & \textbf{4.88} & 4.57 & 4.61 & 4.38 & 4.48 \\
Multi-Turn (Audio) & \textbf{1.25} & 1.08 & 1.04 & 1.21 & 1.23 \\
\AvgDash
\brow \textbf{Avg (Implicit)} & \underline{\textbf{2.78}} & 2.67 & 2.59 & 2.55 & \underline{\textbf{2.78}} \\

\BoldDoubleBot
\end{tabularx}
\end{table*}

\section{Benchmark for End to End Spoken Diglogue models}
\subsection{Task Definition} 
\textbf{WavBench defines the evaluation of end-to-end spoken dialogue models through a tripartite framework: the Pro Subset, the Basic Subset, and the Acoustic Interaction Set.} \textbf{In the Pro Subset}, the task rigorously challenges reasoning-enhanced models with high-difficulty cognitive scenarios, such as multi-step mathematical reasoning and complex coding logic. The model is required to not only achieve factual accuracy but also employ colloquial optimization to simplify intricate logic, thereby ensuring high listenability and reducing cognitive load during auditory information processing. \textbf{In the Basic Subset}, the task establishes a standard for spoken colloquialism in routine interactions. The model is required to prioritize "listenability" through lexical appropriateness, linguistic naturalness, and interactive rapport, strictly distinguishing authentic spoken interaction from rigid text generation. \textbf{In the Acoustic Interaction Set}, evaluation is conducted via explicit instructions and implicit chats. Specifically, explicit instructions cover two dimensions: for understanding, the model must accurately identify significant paralinguistic styles from spoken inputs; for generation, the model must produce spoken responses that strictly adhere to explicit directives. Implicit chats are used to uniformly evaluate understanding and generating abilities, where the model needs to understand the paralinguistic information embedded in spoken inquiries and generate spoken responses that are content-correct and stylistically matched.

\subsection{Evaluation Metrics}
\textbf{In the Colloquial Expression Set}, we leverage Gemini 3 Pro Preview\footnote{\url{https://ai.google.dev/gemini-api/docs/gemini-3?hl=zh-cn}} to implement a hierarchical scoring mechanism for assessing the model's conversational proficiency. A score of 1 is assigned to task failures, defined as instances where the model provides incorrect answers, fails to strictly adhere to instructions, or generates responses to harmful inquiries. For responses that successfully complete the task, we distinguish between scores of 3 and 5 based on conversational naturalness. A score of 5 is awarded to responses that satisfy four specific conversational criteria: (1) Lexical Appropriateness, characterized by the use of everyday lexicon and discourse markers; (2) Linguistic Naturalness, featuring concise and simple sentence structures; (3) Interactive Rapport, involving the use of rhetorical questions and confirmations; and (4) Emotional-Contextual Matching, ensuring the response mirrors natural human communication. Conversely, factually accurate responses that fail to meet these colloquial standards are assigned a score of 3. The corresponding prompt template can be found in Figure~\ref{fig:Code_Eval_Prompt}. \textbf{In the Acoustic Interaction Set}, for the understanding scenario under explicit instructions, we evaluate performance by directly computing the accuracy of the model's predictions against ground-truth labels. For the generation scenario, we employ Gemini 3 Pro Preview to annotate the paralinguistic features of the spoken responses, subsequently calculating accuracy by comparing these annotations with ground-truth labels. In the implicit interaction scenario, which requires joint evaluation of understanding and generation capabilities, we assess performance based on content accuracy and stylistic consistency. Specifically, Gemini 3 Pro Preview is utilized to score the paralinguistic style of the spoken responses, while Gemini 3 Pro Preview evaluates the semantic correctness of the corresponding transcriptions. Both models utilize a scoring scale from 0 to 10. The prompt templates are provided in Figure~\ref{fig:paralinguistic} and Figure~\ref{fig:score_content}.

\subsection{Experimental Results}
We evaluated five end-to-end spoken dialogue models, including \textbf{Qwen3-Omni \cite{xu2025qwen3omni}, Kimi-Audio \cite{kimiaudio}, Mimo-Audio \cite{mimoaudio}, Step-Audio-2-mini \cite{wu2025stepaudio2}, and GPT-4o Audio}\footnote{\url{https://platform.openai.com/docs/models/gpt-4o-audio-preview}}, across five distinct scenarios: \textbf{Colloquial Expression Pro, Colloquial Expression Basic, Explicit Understanding, Explicit Generation and Implicit Chats.} The specific details and configurations of these models are provided in Appendix~\ref{appendix_baseline}.

\subsubsection{Colloquial Expression Pro} 
As shown in Panel A of Table~\ref{tab:wavbench_unified}, the Colloquial Expression Pro subset serves as a rigorous stress test for simplifying intricate logic under high cognitive loads. The results highlight that this subset demonstrates strong discriminative power, effectively distinguishing the reasoning capabilities of different models while revealing significant room for improvement in the field. GPT-4o Audio establishes a distinct lead with an average score of 58.23, yet this figure indicates that even state-of-the-art models are far from mastering natural delivery in complex scenarios. In contrast, open-source models exhibit a marked performance drop, with Qwen3-Omni scoring 39.53 and others like Step-Audio-2 struggling significantly at 30.40. This gap is most pronounced in logic-intensive domains. For instance, in \textbf{Math}, Step-Audio-2 plummets to 22.40, and in \textbf{Logic}, Kimi-Audio scores a mere 26.03, indicating that lightweight models tend to output rigid responses when faced with symbolic reasoning. Even the top-performing GPT-4o Audio drops to 42.60 in Logic, further underscoring the "Cognitive-Acoustic Alignment" gap: current models generally fail to translate complex reasoning into auditorily comprehensible explanations, validating the Pro subset as a challenging benchmark for future reasoning-enhanced audio models.

\subsubsection{Colloquial Expression Basic}
As shown in Panel B of Table~\ref{tab:wavbench_unified}, we report the performance of various spoken dialogue models in the Colloquial Expression Basic subset. We conducted an analysis of the models' ability to maintain conversational liveliness and engagement across routine cognitive tasks. GPT-4o Audio demonstrates dominant performance across all categories, achieving the highest average score of 68.80. It excels particularly in Safety (81.00) and QA (75.60), indicating that large-scale proprietary models have successfully aligned safety guardrails with natural, engaging spoken delivery. Qwen3-Omni follows as a strong contender (55.80), showing a balanced profile with consistent scores above 50 in Creative Writing and Code. Conversely, lightweight models exhibit significant performance disparities across domains: while Step-Audio-2-mini and Mimo-Audio achieve respectable scores in Safety (approx. 60.00-62.00), their performance declines sharply in structured tasks, with Step-Audio-2-mini falling to 30.20 in Math and Mimo-Audio to 33.56 in Instruction. This sharp contrast suggests that while these models can sustain interaction in open-ended chats, they struggle to convert rigid logical constraints or symbolic math into listener-friendly speech, often reverting to mechanical recitation. Overall, the results highlight that bridging the gap between strict logical precision and spoken flexibility remains a critical challenge for the open-source community, necessitating future focus on data that specifically models the "verbalization" of structured knowledge.

\subsubsection{Explicit Understanding}
As shown in Panel C of Table~\ref{tab:wavbench_unified}, we report the accuracy of various end-to-end spoken dialogue models in the explicit understanding scenario. We observe that while models generally achieve high proficiency in Language Identification and Emotion Recognition, they exhibit limited capabilities in distinguishing fine-grained prosodic features such as Pitch, Volume, and Accent, indicating that disentangling specific acoustic attributes remains a significant challenge. Step-Audio-2-mini demonstrates the most robust performance, achieving the highest average score of 57.36\%. It excels particularly in Music and Language, suggesting that its training strategy effectively balances paralinguistic perception with semantic understanding. Kimi-Audio also shows competitive results, performing consistently well across Audio Event Detection and Language, though it struggles significantly with Accent. Surprisingly, while GPT-4o Audio exhibits dominant performance in Language and strong capabilities in Emotion, it underperforms markedly in Age and Accent. This discrepancy highlights that even large-scale proprietary models may prioritize semantic fidelity and emotional alignment over demographic classification tasks. Qwen3-Omni stands out with the highest accuracy in Emotion recognition, yet its ability to classify Gender and Music is relatively weak, indicating an uneven distribution in its feature representation space. Conversely, Mimo-Audio lags behind in most acoustic characteristics, particularly in Audio Event Detection, which could be attributed to a lack of diversity regarding background audio within its training data compared to other spoken dialogue models.

\subsubsection{Explicit Generation}

As shown in Panel D of Table~\ref{tab:wavbench_unified}, we report the accuracy of various spoken dialogue models in the explicit generation scenario. Notably, GPT-4o Audio demonstrates dominant performance across all metrics, particularly excelling in attributes related to speaker identity and emotional state, such as Gender, Emotion, and Age. Qwen3-Omni also exhibits strong capabilities, performing exceptionally well in Emotion and Language, while Step-Audio-2-mini demonstrates competitive results in Language and Music generation. Conversely, models such as Kimi-Audio and Mimo-Audio still lag behind in paralinguistic generation tasks, highlighting their limitations. Specifically, both models struggle significantly with Accent and Audio (background sound) generation, with scores notably lower than their counterparts. This suggests that these models, possibly constrained by the diversity of their acoustic training data, have not yet fully mastered the generation of fine-grained prosodic features and non-speech acoustic events. Additionally, architectural differences appear to impact performance; for instance, Mimo-Audio maintains decent Pitch control despite its lower average score, whereas Kimi-Audio faces challenges across most acoustic characteristics. Currently, the best-performing GPT-4o Audio model achieves an average accuracy of 79.23\%, setting a new benchmark for the field. However, the relatively lower scores across all models in the Background Audio dimension (all below 50\%) indicate that end-to-end dialogue models still have substantial room for improvement in generating complex, realistic acoustic environments beyond pure speech.

\subsubsection{Implicit Dialogue}
As shown in Panel E of Table~\ref{tab:wavbench_unified}, we report the scores of various spoken dialogue models in the implicit scenario. We conducted an analysis of both content and style in spoken responses. In single-turn scenarios, we observed that models generally achieved higher audio scores compared to text scores, with Step-Audio-2-mini reaching the highest audio score of 3.50 despite having the lowest text score. This indicates that models can effectively capture immediate paralinguistic cues, but ensuring semantic precision in short interactions remains challenging. Conversely, in multi-turn dialogue scenarios, the trend reverses: while text scores improved significantly across all models (e.g., Qwen3-Omni reaching 4.88), audio scores declined sharply to the 1.05-1.25 range. This sharp contrast suggests that while audio context helps models maintain semantic coherence (high IQ), maintaining consistent paralinguistic style (EQ) across multiple turns remains a critical bottleneck. Overall, Qwen3-Omni and GPT-4o Audio achieve the highest average score of 2.78, demonstrating the strongest comprehensive ability to process paralinguistic information in implicit interactions.

\section{Conclusion}
In this work, we present WavBench, a comprehensive benchmark designed to evaluate the realistic conversational abilities of end-to-end spoken dialogue models across a tripartite framework comprising cognitive complexity, colloquial delivery, and paralinguistic fidelity. WavBench comprises 17,577 high-quality items totaling 76.5 hours that span five diverse subsets, covering seven cognitive domains and ten paralinguistic attributes to strictly test models in authentic real-world scenarios. We benchmarked five state-of-the-art end-to-end models, including both proprietary giants and emerging open-source systems. Our results demonstrate that WavBench is substantially more challenging than existing benchmarks, particularly in the Pro subset which requires bridging intricate logic with natural spoken expression. Notably, while GPT-4o Audio establishes a distinct lead, open-source models exhibit a marked decline in logic-intensive domains and fine-grained acoustic generation, highlighting a critical "Cognitive-Acoustic Alignment" gap. Further analysis of implicit interactions reveals that maintaining paralinguistic consistency in multi-turn dialogues remains a critical bottleneck despite semantic coherence. We hope WavBench will serve as a rigorous and forward-looking benchmark for advancing robust, reasoning-enhanced spoken dialogue models.

\newpage
\bibliographystyle{neurips_2025}
\bibliography{neurips_2025}


\appendix

\section{ Details about Spoken Dialogue Models}
\label{appendix_baseline}
\begin{itemize}
    \item \textbf{Qwen3-Omni}: A single multimodal model featuring a Thinker-Talker Mixture-of-Experts (MoE) architecture that unifies perception and generation across text, audio, image, and video without performance degradation. It employs a multi-codebook autoregressive scheme for high-fidelity streaming speech synthesis, achieving ultra-low end-to-end latency (234 ms). We utilize the official "Qwen3-Omni-30B-A3B-Instruct" checkpoint in our experiments\footnote{\url{https://github.com/QwenLM/Qwen3-Omni}}.
    
    \item \textbf{Kimi-Audio}: An open-source audio foundation model designed for universal audio understanding, generation, and conversation. It features a novel architecture that combines a 12.5Hz audio tokenizer utilizing both discrete semantic tokens and continuous acoustic features with a flow-matching-based streaming detokenizer. Pretrained on over 13 million hours of diverse audio data, it achieves state-of-the-art performance across speech recognition, audio understanding, and speech conversation tasks. We employ the official "Kimi-Audio-7B-Instruct" checkpoint in our experiments\footnote{\url{https://github.com/MoonshotAI/Kimi-Audio}}.
    
    \item \textbf{MiMo-Audio}: An open-source audio language model that demonstrates strong few-shot learning capabilities by scaling pretraining data to over 100 million hours. It employs a unified decoder-only Transformer architecture with a dual-rate audio tokenization strategy, effectively modeling both low-frame-rate semantic tokens and high-frame-rate acoustic tokens. This design enables versatile capabilities including speech understanding, generation, and complex tasks like voice conversion and style transfer. We utilize the official "MiMo-Audio-7B-Instruct" checkpoint in our experiments\footnote{\url{https://github.com/XiaomiMiMo/MiMo-Audio}}.
    
    \item \textbf{Step-Audio-2}: An industry-strength end-to-end multi-modal large language model designed for advanced audio understanding and speech conversation. It integrates a latent audio encoder with reasoning-centric reinforcement learning to enhance ASR accuracy and paralinguistic responsiveness. Notably, it incorporates discrete audio token generation into language modeling and supports retrieval-augmented generation with external tool use to mitigate hallucinations. Trained on millions of hours of speech and audio data, it achieves state-of-the-art performance on various benchmarks.  We utilize the official "Step-Audio-2-mini" checkpoint in our experiments\footnote{\url{https://github.com/stepfun-ai/Step-Audio2}}.
    
    \item \textbf{GPT-4o Audio}: A proprietary multimodal foundation model developed by OpenAI  that integrates text, audio, and visual processing into a single end-to-end architecture. It excels in real-time speech-to-speech interaction, demonstrating superior capabilities in paralinguistic understanding, logical reasoning, and expressive generation without relying on intermediate transcription. As a closed-source commercial product, we utilize the official API  to assess its performance as a state-of-the-art baseline\footnote{\url{https://platform.openai.com/docs/models/gpt-4o-audio-preview}}.
    
\end{itemize}

\section{Ethical Discussion}
We recognize that assessing vocal attributes such as age, gender, accent and emotion may inadvertently reinforce stereotypes or introduce unfair treatment. For example, systematic misclassification of an accent could disadvantage certain speaker groups in downstream applications. Moreover, using real or synthesized voice recordings without robust consent protocols raises privacy risks and the potential for misuse in voice‑cloning or deepfake generation. The capacity for generating natural, real‑time speech further amplifies these concerns by lowering the barrier for automated impersonation or disinformation campaigns. To address these challenges pragmatically, we perform manual filtering on our audio corpus to remove samples that carry overt biases or sensitive content. Additionally, by open‑sourcing all data curation scripts and evaluation code, we enable researchers and practitioners to audit, reproduce, and extend our methods—encouraging collaborative refinement and helping ensure that spoken dialogue systems built upon our benchmark remain fair and trustworthy.

\begin{figure*}[h]
\centering

\begin{tcolorbox}[
  colback=yellow!6,      
  colframe=yellow!50,    
  width=0.96\linewidth,  
  arc=2pt,               
  boxrule=0.4pt,         
  top=5pt, bottom=5pt, left=8pt, right=8pt, 
  before upper={\setstretch{1.00}\setlength{\parskip}{0.1em}\raggedright} 
]
\footnotesize

\textbf{<System>}

You are a professional designer of single-round conversation instructions tasked with creating single-round emotion training instructions.

\vspace{0.6em}

\begin{enumerate}[leftmargin=*, noitemsep, topsep=0pt]
    \item The user's input reflects the scenario `\{scenario\}' and implies the emotion `\{emotion\}', but must not directly mention emotion words (e.g., ``happy'', ``sad'', ``angry'', etc.).
    \item The model's response should naturally adapt to the implied emotion of the scenario, with a tone close to everyday communication, avoiding imperative or unnatural expressions.
    \item Provide an appropriate emotion label (``\{emotion\}'').
    \item Both the right emotional label must belong to \{EMOTION\_CATEGORIES\}.
    \item User input must be vivid, relevant, and based on the provided scenario.
\end{enumerate}

\vspace{0.6em}

\textbf{Example for `\{emotion\}' emotion:}
\begin{itemize}[leftmargin=*, noitemsep, topsep=0pt]
    \item User: ``User input example for \{emotion\}'',
    \item Model: ``Model response example for \{emotion\}'',
    \item Appropriate label: ``\{emotion\}'',
\end{itemize}

\vspace{0.6em}
Please generate a new single-turn dialogue based on the scenario `\{scenario\}', implying the ``\{emotion\}'' emotion, strictly following the format below:
\begin{verbatim}
User: "User input",
Model: "Model response",
Appropriate label: "{emotion}",
\end{verbatim}

\end{tcolorbox}

\caption{The prompts used to guide the large language model to generate the corpus.}
\label{fig:generate_corpus}
\end{figure*}

\begin{figure}[h]
\centering

\begin{tcolorbox}[
  colback=yellow!6,      
  colframe=yellow!50,    
  width=0.96\linewidth,  
  arc=2pt,               
  boxrule=0.4pt,         
  top=5pt, bottom=5pt, left=8pt, right=8pt,
  before upper={\setstretch{1.00}\setlength{\parskip}{0.1em}\raggedright} 
]
\footnotesize

\textbf{<System>}

\vspace{0.2em}
\textbf{**Important:**} Evaluation must include a ``\textless score\textgreater'' rating.

\vspace{0.2em}

Listen to the response recording and evaluate the \{label\} emotional information in the response recording. 

\vspace{0.6em}

Give an integer score from 1-10 for judging how close the emotional information in the response recording is to \{label\}.

\vspace{0.6em}

    \textbf{**10 points**}: The emotional information in the response recording matches \{label\} exactly.
     
    \textbf{**8 points**}: How close the emotional information in the response to the recording matches \{label\}.
    
    \textbf{**6 points**}: The emotional message of the answer to the recording does not contradict \{label\}.
    
    \textbf{**4 points**}: The emotional message of the response recording is inaccurate for \{label\}.
     
    \textbf{**2 points**}: The emotional information in the recorded response is not \{label\} at all.

\vspace{0.6em}

Responses must be *** in the following format: \texttt{\textless score\textgreater X\textless/score\textgreater} (\textbf{**X is a number between 1 - 10**})

\end{tcolorbox}

\caption{The prompts for guiding large language models to score paralinguistic information.}
\label{fig:paralinguistic}
\end{figure}

\begin{figure}[h]
\centering

\begin{tcolorbox}[
  colback=yellow!6,      
  colframe=yellow!50,    
  width=0.96\linewidth,  
  arc=2pt,               
  boxrule=0.4pt,         
  top=5pt, bottom=5pt, left=8pt, right=8pt,
  before upper={\setstretch{1.00}\setlength{\parskip}{0.1em}\raggedright} 
]
\footnotesize

\textbf{<System>}

\vspace{0.2em}
\textbf{**Important:**} Evaluation must include a ``\textless score\textgreater'' rating.

\vspace{0.2em}

Two paragraphs of text exist below, with the text separated from the text by the \#\#\#\#\# separator. The text after the separator is a response to the text before the separator.You need to give an integer score from 1-10 based on the quality of the answer text content. 
\vspace{0.6em}

You are asked to give a judgment on how reasonable the answer text after the separator is for the text before the separator.

\vspace{0.6em}

    \textbf{**10 points**}: The content of the response text is lively and fits well with the information and emotion of the scene in which the text is asked.
    
    \textbf{**8 points**}: The content of the response fits the scenario of the text, but the response is stiff and there is no clear interaction with the text.
    
    \textbf{**6 points**}: The content of the response text is only relevant to the scenario information in the question text, but does not directly answer the question in a positive way.
    
    \textbf{**4 points**}: The content of the response text is not relevant to the scenario information in the question text, and the scenario information and other constructive factors are completely incompatible.
    
    \textbf{**2 points**}: The content of the response text is not a fluent expression, with grammatical errors and unclear semantics.

\vspace{0.6em}

Responses must be *** in the following format: \texttt{\textless score\textgreater X\textless/score\textgreater} (\textbf{**X is a number between 1 - 10**})

\end{tcolorbox}

\caption{The prompts for guiding large language models to score content information.}
\label{fig:score_content}
\end{figure}

\begin{figure*}[h]
\centering

\begin{tcolorbox}[
  colback=yellow!6,
  colframe=yellow!50,
  width=0.96\linewidth,
  arc=2pt,
  boxrule=0.4pt,
  top=5pt, bottom=5pt, left=8pt, right=8pt,
  before upper={\setstretch{1.00}\setlength{\parskip}{0.1em}\raggedright} 
]
\footnotesize

\textbf{Prompt for Evaluating Spoken Code-Related Dialogues}

You are a professional expert in evaluating the quality of spoken dialog responses for programming/code-related tasks. Your task is to assess the quality of the test model's response by comparing it with the standard reference answer.

\vspace{0.2em}
\textbf{Scoring Criteria (Only use 1, 3, or 5 points):}

\textbf{1 point - Task not completed or incorrect answer:}
\begin{itemize}[leftmargin=*, noitemsep, topsep=0pt]
    \item The test model's response did not correctly answer the programming question in \texttt{spoken\_instruction}.
    \item The answer is incorrect or fails to provide the correct solution/code explanation.
    \item Response is irrelevant to the coding question.
    \item Contains obvious factual errors or logical mistakes in the code.
\end{itemize}

\textbf{3 points - Correct answer but not conversational enough:}
\begin{itemize}[leftmargin=*, noitemsep, topsep=0pt]
    \item Correctly answered the coding question in \texttt{spoken\_instruction} with the right solution/approach.
    \item The explanation matches the technical correctness of \texttt{spoken\_reference}.
    \item However, the expression is stiff, mechanical, or overly formal.
    \item Does not conform to conversational expression habits.
    \item Lacks the feeling of natural human communication.
\end{itemize}

\textbf{5 points - Correct answer and naturally conversational:}
\begin{itemize}[leftmargin=*, noitemsep, topsep=0pt]
    \item Correctly answered the coding question with accurate solution/approach.
    \item The explanation is technically sound and matches \texttt{spoken\_reference} quality.
    \item Expression meets all conversational requirements:
    \begin{itemize}[label=$\bullet$, leftmargin=1.5em, noitemsep, topsep=1pt]
        \item \textbf{(Vocabulary Naturalness)}: Uses everyday, colloquial vocabulary; naturally incorporates tone particles or trendy language; sounds very much like a real person talking.
        \item \textbf{(Syntactic Fluency)}: Primarily uses short, simple sentences that are easy to understand; flexible sentence structures; uses omissions, inversions, and other typical spoken language patterns.
        \item \textbf{(Interactive Rapport)}: Highly interactive and engaging; uses rhetorical questions, confirmations, suggestions to guide the conversation; expresses uncertainty through natural language, making it feel like chatting with a thoughtful partner.
        \item \textbf{(Emotional-Contextual Matching)}: Emotional expression and tone perfectly match the conversation context; conversational style perfectly matches the coding task scenario.
    \end{itemize}
    \item Natural and fluent, sounds like a real person communicating about code.
    \item Expression style is easily acceptable to humans.
\end{itemize}

\vspace{0.2em}
\textbf{Evaluation Process:}
\begin{enumerate}[leftmargin=*, noitemsep, topsep=0pt]
    \item First, determine if the programming question was correctly answered (refer to \texttt{spoken\_reference} to understand the correct technical solution).
    \begin{itemize}[nosep]
        \item If not correct or incorrect answer $\to$ 1 point.
    \end{itemize}
    \item If the answer is correct, then evaluate the conversational level.
    \begin{itemize}[nosep]
        \item Compare with the expression style of \texttt{spoken\_reference}.
        \item Check all 4 conversational criteria.
        \item Stiff and mechanical $\to$ 3 points.
        \item Natural and conversational with all criteria met $\to$ 5 points.
    \end{itemize}
\end{enumerate}

\vspace{0.2em}
\textbf{Important Notes:}
\begin{itemize}[leftmargin=*, noitemsep, topsep=0pt]
    \item \texttt{spoken\_reference} is the standard reference answer, representing the ideal conversational response style for code explanations.
    \item The test model's response does not need to be identical to the reference, but should achieve similar naturalness in tone.
    \item Focus on evaluating whether the answer is technically correct AND whether the expression is natural and conversational.
\end{itemize}

\vspace{0.2em}
\textbf{Output Format:}
Please strictly output the evaluation result in the following JSON format:
\begin{verbatim}
{
    "score": <1 or 3 or 5>,
    "reasoning": "<Detailed reasoning for the score>"
}
\end{verbatim}

\end{tcolorbox}

\caption{Prompt specification for assessing spoken capability in code-related tasks using a hierarchical scoring mechanism.}
\label{fig:Code_Eval_Prompt}
\end{figure*}

\end{document}